# Constraints, Exceptions and Representations

T. Mark Ellison

Centre for Cognitive Science, University of Edinburgh
2 Buccleuch Pl., Edinburgh EH8 9LW, U.K.
marke@cogsci.ed.ac.uk

June 3, 1994

### Abstract

This paper shows that default-based phonologies have the potential to capture morphophonological generalisations which cannot be captured by non-default theories. In achieving this result, I offer a characterisation of Underspecification Theory and Optimality Theory in terms of their methods for ordering defaults. The result means that machine learning techniques for building declarative analyses may not provide an adequate basis for morphophonological analysis.

## 1 Introduction

In other work, I have shown (Ellison 1992, forthcoming) that interesting phonological constraints can be learned despite the presence of exceptions. Each of these constraints imposes a limit the set of possible words at a common level of representation. In this paper, I consider possible limits to the usefulness of these constraints in representing morphemes and finding concise representations of lexical entries.

In order to compare a strictly declarative formalism with other constraint formalisms, a common formal environment must be established. Using model theory to establish the relationship between description and object, and then a modal formalism to define the structures to which constraints apply, we can compare the different effects of strict constraints and defaults. In particular, a strict declarative approach can be compared with other constraint frameworks such as Underspecification Theory (UT) (Archangeli, 1984) and Optimality Theory (OT) (Prince & Smolensky, 1993). This discussion is followed in the latter part of the paper by consideration of the possibility of using machine learning to constraint systems that use defaults.

## 2 Morphophonology

To structure the discussion, I offer four desiderata for morphophonology. The first is that the morphophonology must allow concise lexical representations. Where information is predictable, it should not have to be specified in the lexicon. This desideratum is not a matter of empirical accuracy, rather one of scientific aesthetics. For example, English contains no front rounded vowels, so a vowel which is marked as front in the lexicon need not be marked as unrounded.

The second desideratum is that the morphophonology should allow generalisations to be made over phonologically conditioned allomorphs. For example, a representation of the Turkish plural affixes -lar, -ler, that uses the feature [±front] is superior to a segmental representation because a single representation for the two allomorphs can be achieved by not specifying the value for this feature in the representation of the morph.

The third desideratum requires that the specific allomorphs be recoverable from the generalisations. If -lar and -ler are generalised in a single representation, such as -lAr, then the morphophonology should make the recovery of the allomorphs in the correct environments possible.

The final desideratum is, like the first, a matter of scientific aesthetics: a priori abstractions should not be used in an analysis any more than is necessary. For example, the feature [±front] should not be used in the analysis of a language unless it is motivated by structures in the language itself. This desideratum may conflict with the first: a priori features may result in a more concise representation.

These four desiderata provide a framework for evaluating the relative merits of monostratal systems of phonological constraints with other current theories such as Underspecification Theory and Optimality Theory.

# 3    Model Theory and Modal Logic

A fundamental distinction in any formal account is the distinction between description and object. Failure to make the distinction can lead, at best, to confusion, and, at worst, to paradoxes, such as Russell's Paradox. Because this theory is talking *about* theories, it must make the distinction explicitly by formalising the relationship between description and object. This distinction is pursued in below and developed into a formalism for complex structures in the following section.

## 3.1    Model theory

In model theory, the meaning of a statement in a formal language is provided by means of an INTERPRETATION FUNCTION which maps the statement onto the set of objects for which the statement is true. If $L$ is a language and $W$ is a set of objects, and $\mathcal{P}(W)$ is the set of all subsets of $W$, then the interpretation function $I$ maps $L$ onto $\mathcal{P}(W)$:

$$I : L \to \mathcal{P}(W).$$

As an example, suppose ♣ is a horse, ◇ is a ferret and ♡ is a large stone, and that these are the objects in our world. We might define a language $L_0$ containing the terms *big*, *animate*, *slow* and *human*, and assign these terms the interpretations in (1).

(1)
| Term $T$ | Interpretation $I_0(T)$ |
|---|---|
| *big* | $\{♣, ♡\}$ |
| *animate* | $\{♣, ◇\}$ |
| *slow* | $\{◇, ♡\}$ |
| *human* | $\{\}$ |

This language can be expanded to include the logical operations of conjunction, disjunction and negation. These are provided a semantics by combining the semantics of the terms they apply to.

(2)
| Term | Interpretation |
|---|---|
| $l \in L_0$ | $I_0(l)$ |
| $X \wedge Y$ | $I(X) \cap I(Y)$ |
| $X \vee Y$ | $I(X) \cup I(Y)$ |
| $\neg X$ | $W \setminus I(X)$ |

With this interpretation function, we can determine that *big* $\wedge$ *animate* $\wedge$ *slow* is a CONTRA-DICTION having a null interpretation in $W$, while *big* $\vee$ *slow* is a TAUTOLOGY as $I(big \vee slow)$ is the same as $I(big) \cup I(slow)$ which equals $W$.

The term PREDICATE will be used to describe a statement in a language which has a well-defined interpretation.

## 3.2    Modal logics

Model theory as defined in section 3.1 applies only to domains with atomic, unstructured objects. More complex structures can be captured by extending the theory of models to refer to different worlds and the relationships between them. Such a complex of worlds and relations is called a MODAL logic.

A modal theory consists of a universe $U$ is a set of worlds $W_{j,j \in \mathcal{W}}$, called TYPES, together with a set of relations $R_{k,k \in \mathcal{R}} : W_{\mathbf{dom}(j)} \to W_{\mathbf{cod}(k)}$ from one world to another. Types may contain other types, and whenever a type is so contained, it defines a characteristic relation which selects elements of that subtype from the larger type. A language for this universe is more complex as well, needing a function $\omega : L \to I$ to indicate the type $W_{\omega(l)}$ in which any given expression $l$ is to be interpreted. A MODAL OPERATOR $r_k$ is a special symbol in the language which is interpreted as the relation $R_k$.

Modal operators can combine with predicates to construct new predicates. If $\phi$ is a predicate, $r_k$ is a modal operator and $\omega(\phi) = \mathbf{cod}(k)$ then we can define an interpretation, $I(r_k\phi) \subseteq W_{\mathbf{dom}(k)}$, for $r_k\phi$, namely $R_k^{-1}[I(\phi)]$. Furthermore, we define the type of the expression to be the domain of the functor: $\omega(r_k\phi) = \mathbf{dom}(k)$. The interpretation of any well-formed sentence in this language is a subset of the corresponding world $I(\phi) \subseteq W_{\omega(\phi)}$.

From here on, we will assume that the $R_{k,k \in \mathcal{R}}$ are functions, and call the corresponding operators of the language FUNCTORS. Functors simplify the interpretation of predicates: inverses of functions preserve intersection, so functors distribute over conjunction as well as disjunction.

A path equation defines a predicate which selects entities that have the same result when passed through two different sequences of functions. Suppose that $p$ and $q$ are two sequences of functors with the same first domain and last codomain, and that the composition of the corresponding sequences of functions are $P$ and $Q$ respectively. Then the interpretation of $p = q$ is the set of entities $x$ in the common domain such that $P(x) = Q(x)$.

Suppose the universe $U$ consists of seven worlds, $a$, $b$, $c$, *alphabet*, *nullstring*, *nonnullstring* and *string*. Some of these worlds are built from others: *alphabet* is the disjoint union of $a$, $b$ and $c$, while *string* is the disjoint union of *nullstring* and *nonnullstring*. Linking these types are the three functors shown in (3).

$$(3) \qquad \begin{array}{llll} right & : & nonnullstring & \to & string \\ left & : & nonnullstring & \to & string \\ head & : & nonnullstring & \to & alphabet \end{array}$$

We subject these definitions to the path equation that *right left* $x$ and *left right* $x$ equal $x$ for all non-null strings $x$.

A predicate in the corresponding modal language, using only the characteristic predicates of the types and the functors, might be: *head a* meaning the set of non-null strings whose first letter is **a**, *left head a* $\wedge$ *right head c* to specify the context **a__c**, or *head c* $\wedge$ *right(head a* $\wedge$ *right(head b* $\wedge$ *right null))*.

By the use of functors, we can move from one type to another, or from one item in a type to another item in the same type. Metaphorically, we will call the types joined by functors LOCATIONS, particularly when the type instances are only distinguished by functorial relationships with other types.

In a complex structure, like a string, the functors provide a method for interrogating nearby parts of the the structure within a predicate applying at a given position. By an appropriate choice of types and functors, complex feature structures and/or non-linear representations can be defined. For the sake of simplicity, the discussion in the remainder of this paper will be restricted to strings constructed using the types and functors defined above.

## 3.3 Constraints in a modal theory

In model-theoretic terms, a constraint is any well-formed expression in the language to which an interpretation is attached. Phonologists also use the term, usually intending universal application. It will be used here for a single predicate applying at a particular location in structure.

As an example of a constraint, consider front vowel harmony in Turkish[1]. Informally, we can write this constraint as *if the last vowel was front, so is the current one*. In the format of a phonological rule, this might be written as $[+\text{front}]C^*\underline{V} \Rightarrow [+\text{front}]$, where $C^*$ stands for zero or more consonants. **F** is used to represent the disjunction of all of the front vowels.

---

[1] Turkish has eight vowels, **a**, **e**, **i** the back version of **ı**, **o** and its front correlate **ö**, and **u** and the corresponding front vowel **ü**.

$$
\begin{array}{llll}
(4) & Left & = & (left\ head\ C \wedge left\ Left) \vee \\
& & & left\ head\ F \\
& Constraint & = & head\ F \vee \neg Left
\end{array}
$$

In (4) the left context is abstracted into a named predicate called $Left$. This is because the left context iterates over consonants. This iteration appears in the definition of $Left$ as the recursive call: if the immediate left segment is a consonant, move left and check again. $Left$ succeeds immediately if the immediate left segment is a front vowel.

Note the the predicate defined here imposes no restrictions at all on where it applies except that it be a non-null string. On the other hand, it only applies at the current location in structure. The relationship between constraints and locations is the topic of the next section; first in the discussion of features, and then in the prioritisation of default feature assignment.

# 4    Features, Underspecification and Defaults

# 5    Default Ordering Schemes

# 6    Ordering

Defaults need to be ordered. There are a number of ways that the ordering of groups of defaults can be specified. Three of these are presented here.

## 6.1    Ordering by feature

One method for ordering defaults is to order the features they instantiate. We begin with an ordering on the features, so that, for example, feature [+F] has higher priority than feature [+G], in symbols [+F]≻[+G]. This ordering on features, can then be extended to an ordering on defaults specified with those features.

Suppose $p$ and $q$ are paths in string structure, composed of sequences of $left$ and $right$ functors. Then for any defaults filling in predicates $\delta = p[+F]$ and $\epsilon = q[+G]$, $\delta$ is ordered before $\epsilon$ if and only if [+F] has higher priority than [+G].

Suppose a language is analysed as imposing a higher priority default that front vowels cannot occur after round vowels. Assume that the defaults insert the features [+front] and [+round] in all positions. Given a form $\mathbf{kVtV}$ where $\mathbf{V}$ represents the completely uninstantiated vowel, there are two different instantiations depending on the ordering of the two features.

If the [+front] default applies first, then the resulting form will be $\mathbf{k}[+\text{front}]\mathbf{t}\begin{bmatrix}+\ \text{front} \\ +\ \text{round}\end{bmatrix}$.

If, on the other hand, the [+round] default applies first, the derived form will be $\mathbf{k}[+\text{round}]\mathbf{t}\begin{bmatrix}+\ \text{front} \\ +\ \text{round}\end{bmatrix}$.

## 6.2    Ordering by failure count

Another approach orders defaults instantiating the same feature in different positions. The preferred default minimises the number of contradictions to the default feature value.

Suppose the default feature value to be ordered is [+F]. The failure count default ordering mechanism uses a default predicate for each possible number of exceptions. The predicates, $\delta_i$, are defined in (5).

$$
\begin{array}{llll}
(5) & \delta_i & = & \bigvee_{j+k=i} right\ \grave{\delta}_j \wedge \acute{\delta}_k \\
& \grave{\delta}_0 & = & left\ (null \vee \grave{\delta}_0 \wedge [+\text{F}]) \\
& \grave{\delta}_i & = & left\ (null \vee \grave{\delta}_i \wedge [+\text{F}] \vee \grave{\delta}_{i-1} \wedge [-\text{F}]) \\
& \acute{\delta}_0 & = & right\ (null \vee \acute{\delta}_0 \wedge [+\text{F}]) \\
& \acute{\delta}_i & = & right\ (null \vee \acute{\delta}_i \wedge [+\text{F}] \vee \acute{\delta}_{i-1} \wedge [-\text{F}])
\end{array}
$$

If $\delta_i$ is compatible with a predicate $\phi$, then there is a fully-specified restriction on $\phi$ which has no more than $i$ occurrences of $[-\text{F}]$. The ordering on the defaults is imposed by requiring that for any feature $[+\text{F}_i]$, with the corresponding predicate $\delta_i$, $\delta_i$ has priority over $\delta_j$ iff $i < j$.

Suppose we already have a number of higher priority constraints on stress: that it can only be assigned once and in only one position within a syllable, and that consecutive syllables cannot be stressed. Collapsing the representation of syllables into a single symbol $\sigma$ for convenience, table (6) gives the assignment of stress to a number of partially specified representations. The default feature is [+Stress], and this is applied to minimise the number of failures.

(6)

| φ [±Stress] | | | | | | | | | |
|---|---|---|---|---|---|---|---|---|---|
| After defaults | + | - | + | - | + | - | + | - | - |
| or | - | + | - | + | - | + | - | + | - |
| Location | σ | σ | σ | σ | σ | σ | σ | σ | σ |
| φ [±Stress] | - | | | | | | | | |
| After defaults | + | - | + | - | + | - | + | - | + |
| Location | σ | σ | σ | σ | σ | σ | σ | σ | σ |
| φ [±Stress] | + | | | | | | | | |
| After defaults | - | + | - | + | - | + | - | + | - |
| Location | σ | σ | σ | σ | σ | σ | σ | σ | σ |

## 6.3   Ordering by position

Another possibility is to order defaults by how far away from the starting position they specify their features. There are two simple ways of relating distance to priority: closer means higher priority, or further away means higher priority.

The formal definitions for this kind of default ordering are straightforward. Suppose, once again, that [+F] is the feature value to be filled in by the defaults. Now, $\delta_i$ will denote the specification of a default value at a distance of $i$ functors to the left, or $i$ to the right of the starting position.

(7)
$$\delta_i = right\dot{\delta}_i \wedge \acute{\delta}_i$$
$$\acute{\delta}_0 = [\text{+F}] = \acute{\delta}_0$$
$$\grave{\delta}_{i+1} = left\ \grave{\delta}_i \vee null$$
$$\acute{\delta}_{i+1} = right\ \acute{\delta}_i \vee null$$

To prefer near defaults, prefer $\delta_i$ over $\delta_j$ when $i < j$. For far defaults, do the reverse.

Directional default preferences mimic the application of phonological rules in a left-to-right or right-to-left direction. Using this ordering, directional defaults can restrict some structures which the counting defaults cannot. Consider once again the stress assignments by defaults in table (6). Instead of simply trying to maximise the number of stresses, assume that the starting position is the left end of the word, and that near stresses are given priority. Under this system of defaults, the first of the three underspecified representations is rendered more specific, while the other two make the same restriction. These results are shown in table (8).

(8)

| φ [±Stress] | | | | | | | | | |
|---|---|---|---|---|---|---|---|---|---|
| After defaults | + | - | + | - | + | - | + | - | - |
| Location | σ | σ | σ | σ | σ | σ | σ | σ | σ |
| φ [±Stress] | - | | | | | | | | |
| After defaults | + | - | + | - | + | - | + | - | + |
| Location | σ | σ | σ | σ | σ | σ | σ | σ | σ |
| φ [±Stress] | + | | | | | | | | |
| After defaults | - | + | - | + | - | + | - | + | - |
| Location | σ | σ | σ | σ | σ | σ | σ | σ | σ |

# 7   Three Theories

## 7.1   Underspecification Theory

Within the framework given above, it is possible to define a form of Underspecification Theory. What is described here is not precisely the Underspecification Theory of Archangeli (1984), differing in that the structures described are linear and segmental. This is, however, not a necessary limitation of the framework, and the definition of of underspecification theory presented here could be applied to autosegmental representations if suitable types and functors were defined for them.

In UT, lexical specifications are made in terms of an a priori fixed set of features. For example, Archangeli & Pulleyblank (1989) use the four features [±high], [±low], [±back] and [±ATR] to describe seven Yoruba vowels. All lexical specifications of vowel quality are assumed to involve specifications for some subset of these features.

In the lexical specifications, redundant information is left unmarked. The Yoruba vowel **a** does not need to be marked for any feature other than [+low], because there is only one vowel which is [+low]. Consequently, the feature values [+back], [−high] and [−ATR] are all redundant.

In UT, redundant features are are filled by rule. Special constraints, such as the Redundancy Rule Ordering Constraint (Archangeli, 1984:85) ensure that redundancy rules apply before the features they instantiate are referred to. Furthermore, these constraints apply as often as necessary (Archangeli & Pulleyblank, 1989:209-210). This has the same effect as the automatic specification of redundant feature values in the current framework.

Only one type of feature value is ever lexically specified in UT. Opposite feature values are filled in by default rules. This allows the feature specifications for some segments to be subspecifications of those for other segments.

Apart from the context-free features used in lexical specifications, there are also context-sensitive constraints which are regarded in UT as fully-fledged phonological rules. For example, the Yoruba vowel harmony rule can be summarised as *a vowel on the left of a* [−ATR] *vowel will also be* [−ATR]. Regularity to this constraint in one position may conflict with regularity in another position. In UT, the defaults associated with such constraints are ordered by position: Yoruba vowel harmony applies right-to-left in the sense that constraint applications further from the beginning of the word have higher priority.

This directionality is not the only ordering of defaults. As it happens, there are no [+high] vowels in Yoruba which are also [−ATR]. Consequently, the default rule marking vowels as [+high] can conflict with the default that spreads [−ATR]. In the analysis of Archangeli & Pulleyblank the [+high] default is ordered first. All defaults constructed from the one feature have priority over all defaults built on the other.

The general structure of UT, therefore, is to have an a priori limited set of features for lexical specification and a set of defaults for these features and for constraints. The defaults associated with each feature or constraint are ordered by position.

## 7.2  Optimality Theory

Optimality Theory (Prince & Smolensky, 1993) is apparently a very different theory, but, when classified in terms of its use of defaults, is actually quite similar.

In contrast to UT, OT is deliberately vague about underlying representations. Instead of discussing the manipulation of representations directly, OT refers to their interpretations, terming them CANDIDATE SETS.

Constraints in OT apply exactly like defaults. If they can be imposed without resulting in a contradiction (empty candidate set), then they are. Each constraint imposes a set of defaults, and these are primarily ordered by an extrinsic ordering placed on the constraints. If any two defaults pertaining to two constraints conflict, the default of the higher order constraint is preferred.

As with UT, there is the possibility that the imposition of the the same constraint at different locations will conflict. Rather than ordering these defaults by position, they are ordered by the number of exceptions to the constraint that they allow. If there is a candidate form with a certain number of exceptions, all candidates with more exceptions will be eliminated by the default. This ordering on defaults is the ORDERING BY FAILURE COUNT described earlier.

## 7.3  Exception Theory

In contrast to the other two, more standard, phonological theories, Exception Theory does not use defaults. In ET, each lexical form is fully specified, and any feature in it may be removed so long as this property is preserved.

The set of features includes a feature for each segment type, and a feature for each constraint. While this results in a large set of features, underspecification of redundant features means that many feature specifications may be eliminated. Nevertheless, there will be more feature specifications needed in ET than in, for example, UT, because of the absence of default values.

On the other hand, because ET uses no defaults, there is no need for any form of constraint or rule ordering. All features have an immediate interpretation through the interpretation function, and so a minimum of computation is needed to identify the denotation of a representation.

## 7.4 Summary

Table (9) summarises the attributes of the three theories. UT and OT are primarily distinguished by the use of different methods to order defaults built from constraints. ET differs in that it does not use defaults at all.

(9)

|                  | UT        | OT        | ET |
|------------------|-----------|-----------|----|
| A priori features | √         | ×         | ×  |
| Defaults         | √         | √         | ×  |
| By Feature       | primary   | primary   | ×  |
| By Failure Count | ×         | secondary | ×  |
| By Position      | secondary | ×         | ×  |

# 8 Discussion

Early in this paper, four desiderata for morphophonological theories were introduced. This section considers whether using defaults is advantageous with respect to these desiderata.

## 8.1 Conciseness

The first desideratum sought concise lexical representations for morphemes. Since default-based theories can also exploit underspecification of redundant feature values, they are at least as concise as non-default theories. If there are ever contrastive feature specifications, then they are more concise, allowing one side of the contrast to be left as a default value to be instantiated.

Note that the concept of conciseness which is being used here is feature-counting, not an information-theoretic measure. In a direct application of information theory, contrasting a [+F] feature value with whitespace carries as much information as contrasting it with $[-\mathrm{F}]^2$.

## 8.2 Abstracting and recovering morphemes

Defaults also provide advantages in abstracting morpheme representations from which allomorphs can be recovered. As well as making representations more concise, using defaults allows more allomorphs to be brought together within a single phonological representation. As there are no feature changing rules in the framework, all feature values in the abstract representation must survive to the surface in each allomorph. Conversely, the abstract representation can only contain feature specifications common to all of the allomorphs. So the upper bound on feature specifications for the abstract morpheme is the is the intersection of the featural specifications for all of the allomorphs of the morpheme.

As an example, consider four allomorphs of the Turkish second person plural possessive suffix: -**ınız**, -**iniz**, -**unuz** and -**ünüz**. If the vowels are specified with the three features [±front], [±round] and [±high], then the intersection of the specifications of the four allomorphs is the sequence [+high]**n**[+high]**z**.

While it is always possible to form abstract representations by intersecting feature values (the second desideratum), there is no guarantee that the allomorphs will be readily recoverable (third desideratum). If they are not recoverable, then there is no single featural generalisation which captures the phonological structure of the morphemes.

One important question is whether defaults allow recoverable generalisations about a greater range of morphemes than non-default representations. The answer is yes. If the morphological alternations is one-dimensional, then there is no difference between having defaults and not. Suppose $\delta$ is a default predicate, and, equally, an exception feature. If all allomorphs are specified [$+\delta$] then the abstraction will share this feature, and so the default does not need to apply. Similarly if all allomorphs are specified [$-\delta$], so will the abstract forms be, and the default cannot apply. If the allomorphs vary in their specification for [$\pm\delta$], then the abstraction will not have include a specification for this feature. Consequently, the default will specify $[+\delta]$ when the correct value is $[-\delta]$, and so not fail to produce the correct result. In the non-default interpretation, the representation is never fully specified.

---

[2] It may be possible, nevertheless, to provide an information theoretic basis for the feature-counting notion by couching the feature specifications in a suitable descriptive language.

On the other hand, if the morphological alternations form a two-dimensional paradigm, then it is possible that the paradigm might be decomposable into morphemes only with the use of defaults. Suppose, once again, that $\delta$ is a default predicate and exception feature. The default feature value is $[+\delta]$. Suppose further, that there is a paradigm with the feature specification for $[\pm\delta]$ shown in (10).

(10)

|            | $[-\delta]$ | $[0\delta]$ |
|------------|-------------|-------------|
| $[-\delta]$ | $[-\delta]$ | $[-\delta]$ |
| $[0\delta]$ | $[-\delta]$ | $[+\delta]$ |

The margins show the 'morphemes' extracted by intersecting the feature values. The conjunction of the two $[0\delta]$ specifications is not fully specified for $\delta$, and so its direct interpretation does not recover the corresponding component of the paradigm. If, however, the default $[+\delta]$ is applied, the full specification of the paradigm is recovered.

So it is possible to have paradigms where the morphological components cannot be assigned common phonological representations without the use of defaults[3].

## 8.3   A priori specifications

The final desideratum is the avoidance of a priori information in a model. UT makes use of an a priori set of features for lexical specification. As other generalisations in the formalism are only visible insofar as they affect the values of these features, this limits the possible constraints which can be identified. This is the reason why vowel harmonies such as that of Nez Perce are so problematic for phonologists[4]: the sets of vowels used in the harmony do not have a neat definition in terms of traditional features.

Greater claims about a priori features are made in OT. Prince & Smolensky (1993:3) state that *constraints are essentially universal and of very general formulation ... interlinguistic differences arise from the permutation of constraint-ranking.* In other words, all of the predicates which define features in OT are prior to the analysis of an individual language.

In ET, very little is assumed a priori. Any constraint which captures interesting phonological generalisations about the phonology defines a feature which can be used to specify structure. Because ET does not use defaults, it need not be concerned with ordering constraints, only with finding them. Consequently, interlinguistic differences can only result from distinct sets of constraints.

## 9   Conclusion

In this paper I have presented a rigorous framework for characterising theories that use defaults with phonological structure. The framework provides a straightforward characterisation of Underspecification Theory and Optimality Theory in terms of the action of defaults.

Using this framework, I have shown that non-default theories cannot be sure of capturing all of the generalisations which are available to default theories. For this reason, the non-default constraints learnt by programs such as those described by Ellison (1992, forthcoming), are not as powerful for morphophonological analysis as default-based theories. Furthermore, defaults lead to more concise, and consequently preferable, lexical representations.

The question, therefore, is how to enhance the learning algorithms to involve the use of defaults. The introduction of defaults means that constraints must be ordered; so learning must not only discover the right constraint, it must assign it a priority relative to other constraints. This makes the learning task considerable more complicated. However difficult a solution for this problem is to find, it will be necessary before machine-generated analyses can be sure of competing successfully with man-made analyses.

---

[3] If general predicates are permitted for specifying morphemes, rather than just featural specifications, the distinction between default and non-default systems disappears. If the entries in the paradigm are $\xi_{ij}$, define $\alpha_i$ to be $\bigvee_j \xi_{ij}$ and $\beta_j$ to be $\bigwedge_i (\xi_{ij} \vee \neg\alpha_i)$. Then, so long as the $\alpha_i$ are distinct (which will be the case if the $\xi_{ij}$ are all distinct), then the paradigm will be fully recoverable without defaults.

[4] Anderson & Durand (1988) discuss some of this literature.

# 10    Acknowledgements

This research was funded by the U.K. Science and Engineering Research Council, under grant GR/G-22084 *Computational Phonology: A Constraint-Based Approach*. I am grateful to Richard Sproat and Michael Gasser for their comments on an earlier version of this paper.

# References

Anderson, J. & Durand, J. (1988). Vowel harmony and non-specification in Nez Perce. In H. van der Hulst & N. Smith (Eds.), *Features, Segmental Structure and Harmony Process (Part II)* (pp. 1–17). Foris.

Archangeli, D. (1984). *Underspecification in Yawelmani Phonology and Morphology*. PhD thesis, Massachusetts Institute of Technology.

Archangeli, D. & Pulleyblank, D. (1989). Yoruba vowel harmony. *Linguistic Inquiry, 20*, 173–217.

Asmis, E. (1984). *Epicurus' Scientific Method*. Ithaca, NY: Cornell University Press.

Calder, J. & Bird, S. (1991). Defaults in underspecification phonology. In S. Bird (Ed.), *Declarative Perspectives on Phonology* (pp. 107–125). University of Edinburgh.

Ellison, T. M. (1992). *The Machine Learning of Phonological Structure*. PhD thesis, University of Western Australia, Perth.

Ellison, T. M. (1994). The iterative learning of phonological rules. Technical report, Cognitive Science, University of Edinburgh, Edinburgh.

Prince, A. S. & Smolensky, P. (1993). Optimality Theory: Constraint Interaction in Generative Grammar. Technical Report 2, Center for Cognitive Science, Rutgers University.